\pdfoutput=1

\documentclass[11pt]{article}

\usepackage{acl}

\usepackage{times}
\usepackage{latexsym}
\usepackage{amsfonts}

\usepackage{dsfont}
\usepackage{amsmath,amssymb}
\usepackage{algorithm}
\usepackage{algpseudocode}
\usepackage[T1]{fontenc}

\usepackage[utf8]{inputenc}

\usepackage{graphicx}
\usepackage{subcaption}

\usepackage{microtype}
\usepackage{cinzel} 

\usepackage{multirow}
\usepackage{tabularx}

\usepackage{comment} 
\usepackage{wrapfig} 
\usepackage{calligra}

\usepackage{aurical}

\newcommand{\twoline}[2]{\begin{tabular}[t]{@{}c@{}}#1\\#2\end{tabular}}
\usepackage{algpseudocode}
\usepackage[title]{appendix}

\makeatletter
\def\thickhline{%
  \noalign{\ifnum0=`}\fi\hrule \@height \thickarrayrulewidth \futurelet
   \reserved@a\@xthickhline}
\def\@xthickhline{\ifx\reserved@a\thickhline
               \vskip\doublerulesep
               \vskip-\thickarrayrulewidth
             \fi
      \ifnum0=`{\fi}}
\makeatother

\makeatletter
\newcommand\footnoteref[1]{\protected@xdef\@thefnmark{\ref{#1}}\@footnotemark}
\makeatother

\newlength{\thickarrayrulewidth}
\newcommand*{\affmark}[1][*]{\textsuperscript{#1}}

\newcommand{\DatasetName}{\textit{EuphAug}}
\newcommand{\DatasetNameVOne}{\textit{EuphAug-R}}
\newcommand{\DatasetNameVTwo}{\textit{EuphAug-S}}

\newcommand{\MethodName}{\textsc{{Eureka}}}

\usepackage{amssymb}
\usepackage{pifont}

\usepackage{tipa}
\usepackage{makecell}
\usepackage{enumitem,kantlipsum}

\newcommand{\titlecolor}{violet}
\title{\textcinzelblack{\MethodName}: {\textcolor{\titlecolor} {EU}}phemism
{\textcolor{\titlecolor} R}ecognition
{\textcolor{\titlecolor} E}nhanced Through \\ {\textcolor{\titlecolor} K}NN-based Methods and {\textcolor{\titlecolor} A}ugmentation}

\author{Sedrick Scott Keh\thanks{\quad Equal contribution by S. Keh and R. Bharadwaj}~~\affmark[1], Rohit Bharadwaj\footnotemark[1]~~\affmark[2], Emmy Liu\thanks{\quad Equal contribution by E. Liu and S. Tedeschi}~~\affmark[1], \\ \textbf{Simone Tedeschi\footnotemark[2]~~\affmark[3,4], Varun Gangal \affmark[1], Roberto Navigli \affmark[3]} \\ 
\affmark[1]Carnegie Mellon University, \affmark[2]Mohamed bin Zayed University of Artificial Intelligence,\\ \affmark[3]Sapienza University of Rome, \affmark[4]Babelscape, Italy \\ 
\texttt{\{skeh,mengyan3,vgangal\}@cs.cmu.edu}, \texttt{rohit.bharadwaj@mbzuai.ac.ae} \\ \texttt{\{tedeschi,navigli\}@diag.uniroma1.it}}

\begin{document}
\maketitle

\begin{abstract}
We introduce {\MethodName}, an ensemble-based approach for performing automatic euphemism detection. We (1) identify and correct potentially mislabelled rows in the dataset, (2) curate an expanded corpus called {\DatasetName}, (3) leverage model representations of Potentially Euphemistic Terms (PETs), and (4) explore using representations of semantically close sentences to aid in classification. Using our augmented dataset and kNN-based methods, {\MethodName}\footnote{Our code is available at \url{https://github.com/sedrickkeh/EUREKA}} was able to achieve state-of-the-art results on the public leaderboard of the Euphemism Detection Shared Task, ranking first with a macro F1 score of 0.881.
\end{abstract}

\section{Introduction}
\label{sec:intro}
Euphemisms are mild or indirect expressions used in place of harsher or more direct ones. In everyday speech, euphemisms function as a means to politely discuss taboo or sensitive topics \cite{danescu-niculescu-mizil-etal-2013-computational}, to downplay certain situations \cite{Karam2011TruthsAE}, or to mask intent \cite{magu-luo-2018-determining}. 
The Euphemism Detection task is a key stepping stone to developing natural language systems that are able to process \cite{tedeschi-etal-2022-id10m, liu-etal-2022-testing, jhamtani2021investigating} and generate non-literal texts. 

In this paper, we detail our methods to the Euphemism Detection Shared Task at the EMNLP 2022 FigLang Workshop\footnote{\url{https://sites.google.com/view/figlang2022/home?authuser=0}}. We achieve performance improvements on two fronts:

\begin{enumerate}[wide, labelwidth=!, labelindent=0pt]
    \item \textbf{Data} -- We explore various data cleaning and data augmentation \cite{shorten2019survey,feng2021survey,dhole2021nl} strategies. We identify and correct potentially mislabelled rows, and we curate a new dataset called {\DatasetName} by extracting sentences from a large unlabelled corpus using semantic representations of the sentences or euphemistic terms in the initial training corpus. 
    \item \textbf{Modelling} -- We explore various representational and design choices, such as leveraging the LM representations of the tokens for euphemistic expressions (rather than the \texttt{[CLS]} token) and incorporating sentential context through kNN augmentation and deep averaging networks.
\end{enumerate}

Using these methods, we develop a system called {\MethodName} which achieves a macro F1 score of 0.881 on the public leaderboard and ranks first among all submissions. We found the data innovations to be more significant in our case, indicating that euphemistic terms can be classified with some accuracy if potentially euphemistic spans are identified earlier in the pipeline. 

\section{Task Settings and Dataset}
\label{sec:dataset}
\begin{table*}[t]
    \small
    \centering
    \begin{tabular}{p{8cm}|c|c|c|c}
         \textbf{Sentence Containing PET} & \twoline{\textbf{Sense}}{\textbf{(Euph.)}} & \twoline{\textbf{Sense}}{\textbf{(Non-Euph.)}} & \twoline{\textbf{Label}}{\textbf{(Original)}} & \twoline{\textbf{Label}}{\textbf{(Corrected)}} \\
         \hline \hline
         Does your software collect any information about me, my listening or my surfing habits? Can it be \textcolor{red}{<disabled>}? & Handicapped & Switched off & 1 & 0 \\
         \hline
         Europe developed rapidly [...] Effective and \textcolor{red}{<economical>} movement of goods was no longer a maritime monopoly. & \twoline{Prudent or}{frugal} & \twoline{Related to}{the economy} & 0 & 1 \\
         \hline
         The Lancers continued to hang on to the \textcolor{red}{<slim>} one-point line as Golden West started a possession following [...] & \twoline{Thin (physical}{appearance)} & \twoline{Thin (non-}{physical)} & 1 & 0 \\
    \end{tabular}
    \caption{Examples of incorrectly labelled sentences identified by our data cleaning pipeline. The label is 1 if the term is used euphemistically, 0 otherwise.}
    \label{tab:incorrect-labels}
\end{table*}

\subsection{Task Settings}
The task and dataset are specified by the Euphemism Detection Shared Task, which uses a subset of the euphemism detection dataset of \citet{cats-pets}. The goal of the task is to classify a Potentially Euphemistic Term (PET) enclosed within delimiter tokens as either literal or euphemistic in that context. The training set contained 207 unique PETs and 1571 samples, of which 1106 are classified as euphemisms.

\subsection{Data Cleaning}
\label{subsec:data-cleaning}
\citet{cats-pets} characterize common sources of ambiguity and disagreement among annotators. However, while exploring the data, we also spotted some rows which were, beyond a reasonable doubt, mislabelled (\autoref{tab:incorrect-labels}). This is an artifact of many human-annotated datasets \cite{frenay-label-noise} and is a potential source of noise that could negatively affect performance \cite{Nazari2018EvaluationOC}. 

\begin{figure*}[ht!]
    \centering
    \includegraphics[width=\textwidth]{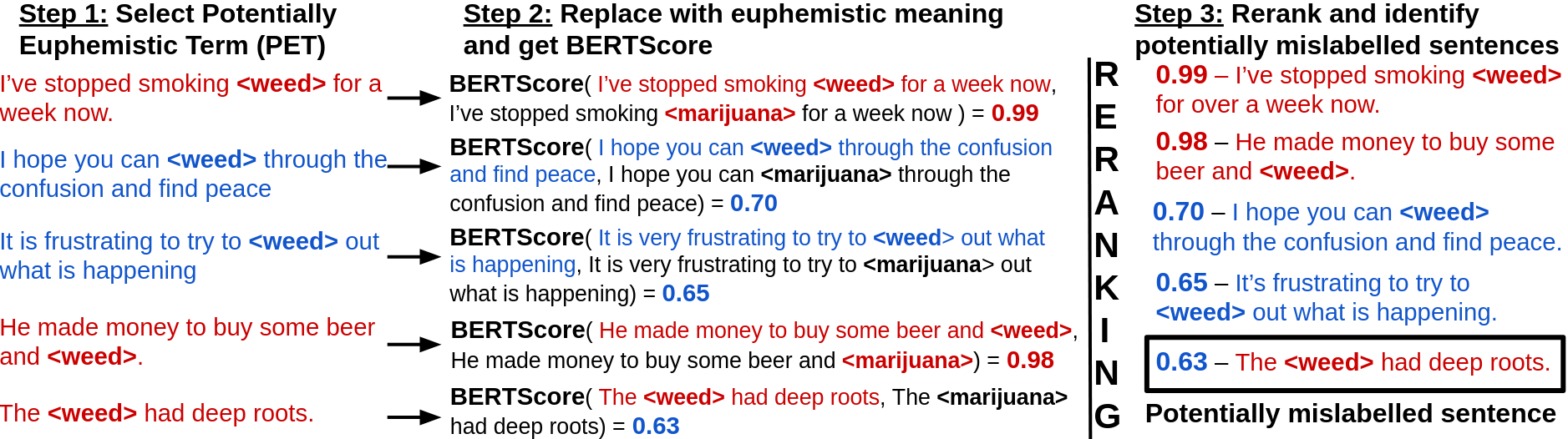}
    \caption{Example of our data cleaning pipeline to automatically identify potentially mislabelled sentences. Red indicates positively classified sentences and blue indicates negatively classified sentences.}
    \label{fig:data-cleaning-pipeline}
\end{figure*}

Motivated by this, we design a data cleaning pipeline to quickly identify and correct such errors (\autoref{fig:data-cleaning-pipeline}). Since the goal is simply to correct as many errors as possible (rather than to be perfectly accurate), we take a few heuristic liberties in our design choices. First, to maximize yield and avoid dealing with less impactful PETs, we filter out PETs which appear $<$10 times or are classified as positive/negative $>$80\% of the time. This leaves us with 33 PETs. We then manually curate a sense inventory (euphemistic vs. non-euphemistic senses) using context clues and BabelNet definitions \cite[v5.0]{navigli2012babelnet}. To ensure the quality of the sense inventory, we have multiple members of our team look through the assigned euphemistic and non-euphemistic senses and verify their appropriateness. Next, for each sentence, we replace the PET with its euphemistic meaning and calculate the BERTScores \cite{bert-score} between the initial sentences and PET-replaced sentences. Replacing euphemistic PETs should not change the semantics drastically and hence should result in a high BERTScore, while replacing non-euphemistic PETs would lead to a low BERTScore. To identify potentially misclassified sentences, we therefore look for positively-classified sentences with low BERTScores or negatively-classified sentences with high BERTScores. We heuristically set this threshold at the halfway mark: if a sentence is among the top half of BERTScores and has a negative label (or among the bottom half and has a positive label), then we flag it as "potentially mislabelled". We end up with 203 potentially mislabelled sentences.

Once these potentially mislabelled sentences have been identified, we go through them manually and correct the ones which we identify as incorrectly labelled, such as the ones in \autoref{tab:incorrect-labels}. In cases where we are unsure of what the label should be (e.g. ambiguous cases as mentioned in \citet{cats-pets}), we leave the original label. As was done with the sense inventories, multiple members of our team then verify that the corrections made are appropriate. Although this still involves some human labor, it is much more tractable as compared to having to go through the entire dataset. Out of the 203 potentially mislabelled rows, we modify the labels of 25 of them.


\subsection{{\DatasetName} Corpus}
\label{subsec:dataset}
In addition to data cleaning, we also use data augmentation techniques to gather an extended corpus, which we call {\DatasetName}. We explore two variants of {\DatasetName}, as outlined below:
\begin{enumerate}[wide, labelwidth=!, labelindent=0pt]
    \item \textbf{Representation-Based Augmentation} -- 
    We search in an external corpus for additional sentences in which specific PETs appear, then assign a label to these PETs based on their vector representations. We call this procedure {\DatasetNameVOne}.
    
    Let our training set (provided by task organizers) be $S$. Consider a PET $p$, which appears in sentences $s_1, s_2, \dots s_k \in S$, with corresponding labels $l_{s_1}, l_{s_2}, \dots l_{s_k} \in \{0,1\}$.
    We search in an external corpus $C$ (i.e., WikiText) for $n$ sentences $c_1, \dots, c_n$ containing the PET $p$. Finally, for each sentence $c_1, …, c_n$ we assign label $l_{c_j}$ as follows:
    
    \begin{algorithm}
    \caption{EuphAug-R}\label{alg:euph}
    \begin{algorithmic}
    \small
    \State \textbf{Task:} Given sentence $c_j$ containing PET $p$, assign $l_{c_j}$.
    \State \textbf{for} {$s_i \in \{s_1, s_2, \dots s_k\}$} \textbf{do}
    \State \hspace{\algorithmicindent} Find ~$\text{dist}_i = \text{dist}(s_i, c_j)$
    \State Find $M = \arg\max\{\text{dist}_1, \text{dist}_2, \dots, \text{dist}_k\}$.
    \State Find $m = \arg\min\{\text{dist}_1, \text{dist}_2, \dots, \text{dist}_k\}$.
    \If{dist$_{M}$ $\geq \, \delta \, \land \, $ ($\lvert$ dist$_{M}$ - $\delta$ $\rvert$ $>$ $\lvert$ dist$_{m}$ - $\epsilon$ $\rvert$)} 
        \State Add $c_j$ to augmented corpus with label $l_{c_j} = l_{s_M}$
    \ElsIf{dist$_{m}$ $\leq \, \epsilon \, \land \, $ ($\lvert$ dist$_{m}$ - $\epsilon$ $\rvert$ $>$ $\lvert$ dist$_{M}$ - $\delta$ $\rvert$)}
        \State Add $c_j$ to augmented corpus with label $l_{c_j} = 1-l_{s_M}$
    \Else
        \State Do not add $c_j$ to augmented corpus.
    \EndIf 
    \end{algorithmic}
    \end{algorithm}

    \noindent where $\delta$ and $\epsilon$ are manually-tuned thresholds, and dist($a, b$) represents the cosine distance between the sentential embeddings \footnote{\url{https://www.sbert.net/}} of $a$ and $b$.
    In other words, we augment our corpus with a sentence $c_j$ only if it is sufficiently similar to, or sufficiently different from, all sentences containing the PET in $S$. We set $n=20$ as the maximum number of sentences extracted from $C$ for a PET $p$, and obtain a corpus of around $4700$ additional examples.
    
    \item \textbf{Sense-Based Augmentation} -- While {\DatasetNameVOne} aims to augment the dataset by finding existing sentences which already contain the PETs, this sense-based approach, instead, considers sentences which contain the senses of the PETs. This is done using the sense inventories created in Section \ref{subsec:data-cleaning} and searching the WikiText corpus. For instance, to find new sentences containing "disabled", we do not search directly for appearances of ``disabled''. Rather, we search for instances of ``handicapped'' and replace these occurrences with ``disabled'' to obtain our positive examples. We then search for instances of ``switched off'' and replace those occurrences with ``disabled'' to obtain our negative examples. We call this expanded corpus {\DatasetNameVTwo}. We sample at most 20 new sentences for each sense (if there are less than 20 occurrences in WikiText, we take all of those present). In addition, some words have senses which cannot be summarized concisely in a single expression (e.g. ``slim'' in Table \ref{tab:incorrect-labels}), so we drop these from our search terms. The final {\DatasetNameVTwo} contains 950 rows.
\end{enumerate}

\section{Methodology}
\label{sec:methodology}
\begin{table*}[t]
    \small
    \centering
    \begin{tabular}{l|l|l|c|c|c}
        \textbf{Feature Tested} & \textbf{Model} & \textbf{Dataset} & \textbf{P} & \textbf{R} & \textbf{F1} \\
        \hline \hline 
        - & RoBERTa-large & Original & 0.8756 & 0.8168 & 0.8399 \\
        \hline
        1) Data Cleaning & RoBERTa-large & Cleaned & 0.8617 & 0.8300 & 0.8435 \\
        \hline
        \multirow{2}{*}{2) Data Augmentation} & RoBERTa-large & Original+{\DatasetNameVOne} & 0.8529 & 0.8388 & 0.8452 \\
        & RoBERTa-large & Original+{\DatasetNameVTwo} & 0.8728 & 0.8306 & 0.8481 \\
        \hline
        3) PET Embedding & RoBERTa-large+PET & Original & 0.8694 & 0.8408 & 0.8533 \\
        \hline
        \multirow{2}{*}{4) Additional Context} & RoBERTa-large+KNN & Original & 0.8769 & 0.8210 & 0.8411 \\
        & RoBERTa-large+DAN & Original & 0.8481 & 0.7983 & 0.8181 \\
        \hline
        \multirow{3}{*}{Final Models} & RoBERTa-large+PET & Cleaned & 0.8728 & 0.8471 & 0.8582 \\
        & RoBERTa-large+PET & Cleaned+{\DatasetNameVTwo} & 0.8692 & 0.8584 & 0.8633 \\
        & RoBERTa-large+PET+KNN & Cleaned & 0.8792 & 0.8517 & 0.8635 \\
        \hline
        Final Ensemble & Model 1 + Model 2 + Model 3 & - & \textbf{0.8994} & \textbf{0.8788} & \textbf{0.8884} \\
    \end{tabular}
    \caption{We independently test 4 features. The final models leverage one or more of these features, and the final ensemble combines the 3 final models. Results are averaged over 10 random seeds. For the final ensemble, since we are just interested in the best possible model, we pick the best random seeds from each model instead of averaging. The three best seeds have F1-scores of 0.8734, 0.8864, and 0.8842, which is slightly improved by our ensembling.}
    \label{tab:main-results}
\end{table*}

For our baseline model, we use a pretrained RoBERTa-large model \cite{liu2019roberta}. For evaluation, we use the macro F1-score, as specified by the shared task description.

\subsection{PET Embeddings}
\label{subsec:pet}
We leverage the embeddings of PET expressions. While models usually perform classification by passing the \texttt{[CLS]} token embedding to a final classifier layer, we instead pass the embeddings of PET tokens. If there are multiple tokens within the PET, we take the sum of these tokens.  
We hypothesize that \texttt{[CLS]} embeddings lose out on the discriminatory power due to pooling of all the embeddings in a sentence, and that using the PET embeddings as signals can better allow the model to focus specifically on the PET senses.

\subsection{Making use of context}
\label{subsec:KNN}

Additionally, we explore using context outside of the PET embeddings. Intuitively, euphemistic and non-euphemistic terms tend to be used in slightly different contexts, with euphemistic terms often being used to discuss sensitive topics. We experiment with two ways to make use of this additional context, as detailed below.

\subsubsection{kNN Augmentation}

Inspired by work on retrieval-based language models \cite{alon2022neuro, knnlm}, we augment the baseline model with a kNN store of the training set, and interpolate the classification probabilities of the base model and a kNN-based model. We follow the usual setup for such a model, with the exception that $y$ is a binary variable indicating euphemistic/non-euphemistic rather than a token from the vocabulary. 

In \autoref{eq:knn}, $\mathcal{N}$ refers to the 5 closest neighbours to $x$ in the training set retrieved through cosine similarity with the \texttt{[CLS]} token generated by RoBERTa, or $f(x)$. $(k_i, v_i)$ refers to the key and value, in this case \texttt{[CLS]} tokens for other sentences, and a binary variable, respectively. In Equation \ref{eq:knn-2}, this value is combined with the probabilities from the base PET model.
\begin{equation}
    \label{eq:knn}
    \scriptstyle
    p_{\text{kNN}}(y | x) ~\propto~ \sum_{(k_i, v_i) \in \mathcal{N}} \mathds{1}(y_i = v_i) \exp{(-d(k_i, f(x)))}
\end{equation}
\begin{equation}
    \scriptstyle
    \label{eq:knn-2}
    p(y|x) = \lambda p_{kNN}(y|x) + (1 - \lambda) p_{PET}(y|x)
\end{equation}
\subsubsection{Deep Averaging Network}

Additionally, we experiment with a Deep Averaging Network (DAN) over embeddings for all the tokens of the sentence \cite{iyyer-etal-2015-deep}. 
For a sentence with tokens $x_1, ..., x_N$, we take the mean vector for the entire sentence: $z = \frac{1}{N} \sum_{i = 1}^{N} x_i$.
We then pass the mean vector through a linear layer with dropout before a second linear layer which outputs to $\mathbb{R}^2$. Note that unlike the original DAN, we do not drop out tokens, as this was found to hurt performance in preliminary experiments.

\subsection{Ensembling}
\label{subsec:ensembling}
\begin{figure}
    \centering
    \includegraphics[width=0.9\columnwidth]{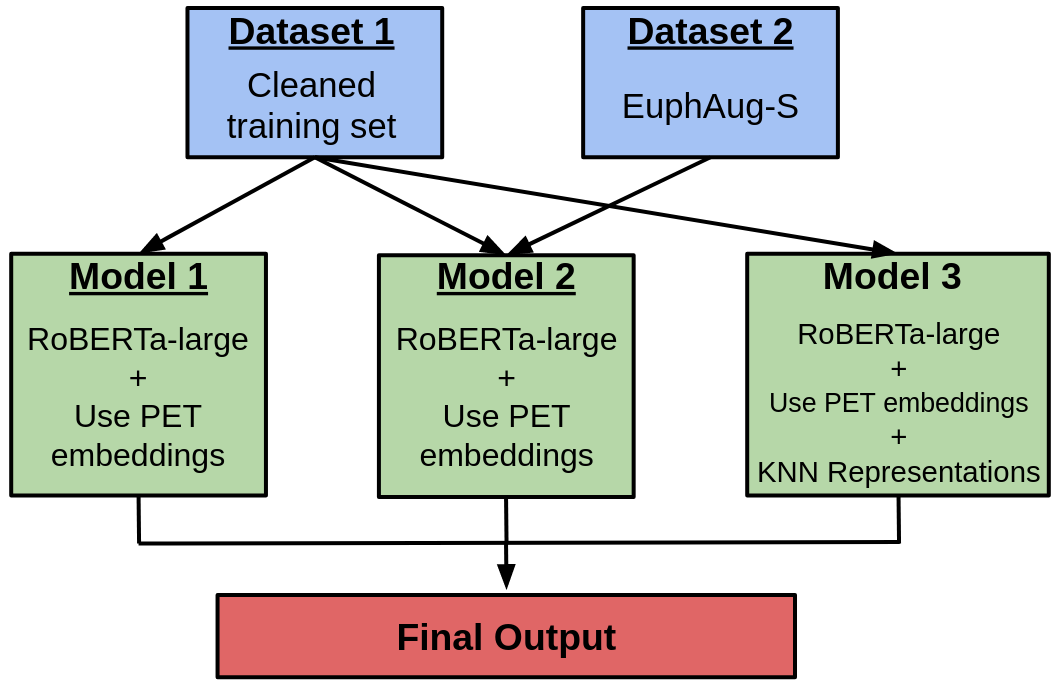}
    \caption{Models and datasets used in the ensemble.}
    \label{fig:models}
\end{figure}
Our final model consists of an ensemble of 3 different models, as shown in Figure \ref{fig:models} and Table \ref{tab:main-results}. For this ensemble, we simply consider a majority vote of the outputs of the 3 models.

\section{Experiments and Results}
\label{sec:experiments_and_results}

\subsection{Implementation Settings}
\label{sec:implementation-settings}

We split our dataset into train-validation-test splits with an 80-10-10 ratio. Note that this splitting is done before any data cleaning or augmentation, so the validation and test sets are not affected by these processes.
Further implementation details are provided in Appendix \ref{appendix:implementation-settings}.

\subsection{Automatic Evaluation Results}
\label{subsec:auto-eval-results}
The main results are shown in Table \ref{tab:main-results}. We independently test 4 features, namely data cleaning, data augmentation, PET embedding, and kNN. Based on the results of these tests, our final models then use combinations of some or all of these features. From the results in \autoref{tab:main-results}, we make the following observations:
\begin{enumerate}[wide, labelwidth=!, labelindent=0pt]
\item \textbf{The data augmentation methods lead to slight increases in performance.} This is true for both data cleaning and augmentation, demonstrating the usefulness of reducing noise and adding high-quality training data. In general, augmentation methods lead to larger gains because adding more data is especially useful in our task, where each PET may appear in the original training data only a few times. 

\item \textbf{Using embeddings of the PET embeddings (instead of the \texttt{[CLS]} classifier token) significantly increases performance.} As hypothesized, this is likely because the \texttt{[CLS]} token may have too wide of a scope since it needs to represent the entire sentence, while the PET tokens can specifically give us information about the PET terms we are trying to classify.
\item \textbf{KNN models lead to slight increase, while DAN models lead to significant decrease, in performance.} In general, our changes in the data side have much greater effects than our changes in the modelling side. For kNN, we think that the neighbors may provide slight signals but are likely drowned out by the original logits, which leads to incremental changes.

We note that the advantages of the kNN method may increase with more data, as this method benefits greatly from a larger datastore. However, as \textit{EuphAug-S} has a relatively large number of examples compared to the training data, we decided to construct the datastore based on only the original training data, as we did not know if there was any significant domain shift between the test data and \textit{EuphAug-S}, and we did not judge the additional samples to be worth this risk.

\end{enumerate}
These three observations motivate our choices for final models to ensemble. In addition, we submit our final ensembled model to the Shared Task leaderboard, and it received an F1 score of 0.881, ranking first place among all submissions.

\section{Related Work}
\label{sec:related_work}
Euphemism detection is a relatively underexplored task. In this paper, we use the euphemism PET dataset gathered by \citet{cats-pets}. \citet{lee-etal-2022-searching} also use this dataset, but for the task of extracting PETs from a given sentence. In the past, other methods have focused specifically on certain types of euphemisms, such as drugs \cite{zhu2021selfsupervised}, firing/lying/stealing \cite{felt-riloff-2020-recognizing}, and hate speech \cite{magu-luo-2018-determining}.

Below, we further detail some of the methods and techniques previously explored in this area. \citet{zhu2021selfsupervised} use BERT and the masked language model objective to create candidate euphemisms based on input target keywords of sensitive topics. \citet{zhu-bhat-2021-euphemistic-phrase} extend this to multi-word euphemistic phrases using SpanBERT. Similar to the previous paper, they also generate and filter a list of euphemistic phrase candidates, then rank these candidates using probabilities from the masked language model. Meanwhile, \citet{felt-riloff-2020-recognizing} use sentiment analysis methods to detect euphemisms, exploring various properties associated with sentiment such as affective polarity, connotation, and intensity.

Other studies contextualize euphemism detection in a specific use case. For instance, \citet{magu-luo-2018-determining} train models to detect hateful content or euphemistic hate speech. They employ word embeddings and network analysis, creating clusters of euphemisms by using eigenvector centralities as a ranking metric. Furthermore, euphemism detection can also be used in crime detection. \citet{Yuan2018ReadingTC} analyze jargon from cybercrime marketplaces to find patterns in phrases or code words commonly used in underground communications. However, these two methods use static word embeddings, which do not take into account the context. This may affect performance, as context is very important for euphemisms. In contrast, our method uses context-aware embeddings of transformer-based models.

\section{Conclusion and Future Work}
\label{sec:conclusion_future_work}
We proposed {\MethodName}, a method for classifying euphemistic usage in a sentence. This is an ensemble model that uses ideas such as data cleaning, data augmentation, representations of Potentially Euphemistic Terms (PETs), and k-nearest-neighbor predictions. Our {\MethodName} system achieves a score of 0.881 and ranks first on the public leaderboard for the Shared Task.

In the future, we hope to extend our methods to dysphemisms or other figurative language instances. It is also interesting to consider a zero-shot setting for euphemism detection, where euphemisms during test time are unseen during training. Figurative language generation, rather than detection, could also be a fruitful area to explore.

\section*{Limitations}
Our current model and classifer are deficient in terms of their interpretability on certain aspects, and it would be interesting to explore more interpretable models to ensure that the features used to classify euphemisms can transfer to other scenarios. The models and datasets are limited to English \cite{bender2018data}, and euphemisms in other languages are definitely worth exploring. However, this was not in the scope of the shared task

Due to computational resources, we were not able to explore larger models. For example, it is possible that larger models such as GPT-J or GPT-Neo would perform better on this task. However, we leave this to future work.

\section*{Acknowledgments}
\begin{center}
\noindent
    \begin{minipage}{0.1\linewidth}
        \begin{center}
            \includegraphics[scale=0.2]{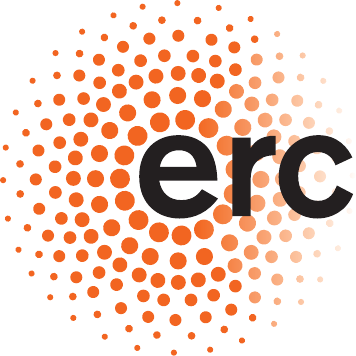}
        \end{center}
    \end{minipage}
    \hspace{0.01\linewidth}
    \begin{minipage}{0.70\linewidth}
        The authors gratefully acknowledge the support of the ERC Consolidator Grant MOUSSE No.\ 726487 under the European Union's Horizon 2020 research and innovation programme.
    \end{minipage}
    \hspace{0.01\linewidth}
    \begin{minipage}{0.1\linewidth}
        \begin{center}
            \includegraphics[scale=0.08]{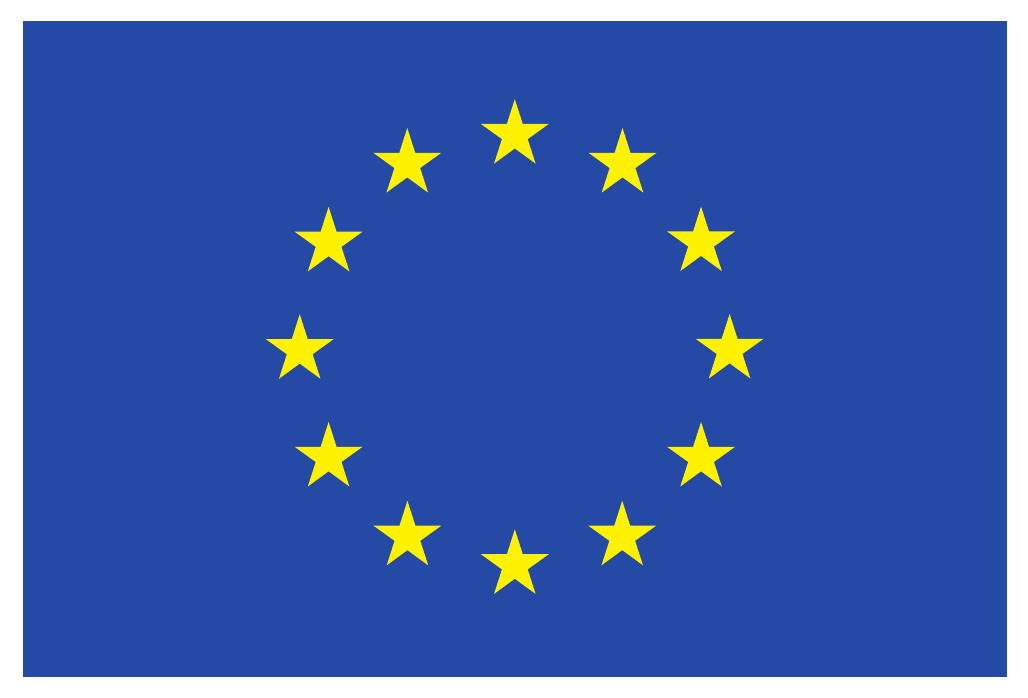}
        \end{center}
    \end{minipage}\\
    
\end{center}

    Thanks to Graham Neubig for discussion and helpful suggestions.

\bibliography{anthology,custom}
\bibliographystyle{acl_natbib}

\newpage
\appendix
\begin{appendices}

\section{Implementation Settings}
\label{appendix:implementation-settings}
For most methods, we use a batch size of 4, learning rate of 5e-6, and we train for 10 epochs. Training was done mostly on a Google Colaboratory environment using Tesla V100, P100 GPUS, and on a workstation having NVIDIA Quadro RTX 6000 with 24GB of VRAM. With RoBERTa-large, training for 10 epochs took around 30-40 minutes. We use the HuggingFace library \cite{wolf-etal-2020-transformers} for model implementation, as well as for implementing the Trainer function. All other hyperparameters (e.g. learning rate decay, warmup steps, etc.) follow the default ones used by the Trainer function in HuggingFace.

\end{appendices}

\end{document}